\documentclass{article}
\usepackage{spconf,amsmath,epsfig}
\usepackage{spconf,amsmath,graphicx,spconf,amsmath,graphicx,epstopdf,diagbox,amsfonts}
\usepackage{times}
\usepackage{epsfig}
\usepackage{amssymb}
\usepackage{multirow}
\usepackage{subfigure}
\usepackage{mathtools}
\begin{document}\sloppy

% Example definitions.
% --------------------
\def\x{{\mathbf x}}
\def\L{{\cal L}}

% Title.
% ------
\title{Incorporating Intra-Class Variance to Fine-Grained Visual Recognition}
%
% Single address.
% ---------------
\name{YAN BAI$^{*,1,2}$, FENG GAO$^{*,2}$, YIHANG LOU$^{1,2}$, SHIQI WANG$^{3}$, TIEJUN HUANG$^{2}$, LING-YU DUAN$^{2}$ \thanks{$^{*}$YAN BAI and FENG GAO are joint first authors. }\thanks{LING-YU DUAN is the corresponding author.}}

\address{$^{1}$SECE of Shenzhen Graduate School, Peking University, Shenzhen, China \\
$^{2}$National Engineering Lab for Video Technology, Peking University, Beijing, China \\
$^{3}$Rapid-Rich Object Search Laboratory, Nanyang Technological University, Singapore\\}

%\name{Anonymous ICME submission}
%\address{}

\maketitle

\begin{abstract}
Fine-grained visual recognition aims to capture discriminative characteristics amongst visually similar categories. The state-of-the-art research work has significantly improved the fine-grained recognition performance by deep metric learning using triplet network. However, the impact of intra-category variance on the performance of recognition and robust feature representation has not been well studied. In this paper, we propose to leverage intra-class variance in metric learning of triplet network to improve the performance of fine-grained recognition. Through partitioning training images within each category into a few groups, we form the triplet samples across different categories as well as different groups, which is called Group Sensitive TRiplet Sampling (GS-TRS). Accordingly, the triplet loss function is strengthened by incorporating intra-class variance with GS-TRS, which may contribute to the optimization objective of triplet network. Extensive experiments over benchmark datasets CompCar and VehicleID show that the proposed GS-TRS has significantly outperformed state-of-the-art approaches in both classification and retrieval tasks.
\end{abstract}
\section{Introduction}
\label{sec:intro}
Fine-grained visual recognition aims to reliably differentiate fine details amongst visually similar categories. For example, fine-grained car recognition \cite{yang2015large,krause20133d} is to identify a specific car model in an image, such as ``Audi A6 2015 model''. Recently, more research efforts in fine-grained visual recognition have been extended to a variety of vertical domains, such as recognizing the breeds of animals \cite{parkhi2012cats, khosla2011novel, berg2014birdsnap},
the identities of pedestrians \cite{ahmed2015improved, yi2014deep, ding2015deep} and the types of plants \cite{qian2015fine, cui2015fine, rejeb2013vantage}, etc. The challenges of fine-grained visual recognition basically relate to two aspects: inter-class similarity and intra-class variance. On the one hand, the instances of different fine categories may exhibit highly similar appearance features. On the other hand, the instances within a fine category may produce significantly variant appearance from different viewpoints, poses, motions and lighting conditions. %(\textit{i.e.}, intra-class variance).  %when directly applying CNN to differentiate the subtle inter-class differences (e.g. similar appearance or attributes). One difficulty lies in the large intra-class variations (e.g. different viewpoints, poses, motions and light conditions), which may bring difficulties in learning discriminative feature.

To mitigate the negative impact of inter-class similarity and/or intra-class variance on the fine-grained visual recognition, lots of research work has been done \cite{xiao2015application, zhang2014part,weinberger2009distance}. Various part-based approaches \cite{xiao2015application, zhang2014part} have been proposed to capture the subtle ``local" structure for distinguishing classes and reducing the intra-class variance of appearance features from the changes of viewpoint or pose, etc. For example, for fine-grained birds recognition in \cite{zhang2014part}, % head and body two parts are separately,leveraged to enhance feature representation.
zhang \textit{et al}. proposed to learn the appearance models of parts (\textit{i.e.}, head and body) and enforce geometric constraints between parts. However, part-based methods rely on accurate part localization, which would fail in the presence of large viewpoints variations. In addition, recently, more promising methods \cite{weinberger2009distance, wang2014learning, schroff2015facenet} based on metric learning, which aims to maximize inter-class similarity distance and meanwhile minimize intra-class similarity distance, have been proposed. In particular, a sort of triplet constraint in \cite{weinberger2009distance} is introduced to learn a useful triplet embedding based on similarity triplets of the form ``sample $A$ is more similar to sample $P$ in the same class as sample $A$ than to sample $N$ in a different class".

%Deep metric learning that aims to adjust samples distance by constraining the same class samples together and pushing the different away in feature space, such as the triplet network \cite{weinberger2009distance} which has achieved state-of-the-art performance in both pedestrian re-identification \cite{ding2015deep} and face recognition \cite{schroff2015facenet}.
On the other hand, some methods \cite{zhang2015embedding,zhou2015fine} utilize multiple labels, which are meant to denote the intrinsic relationship of properties in images, to learn a variety of similarity distances of relative, sharing or hierarchical attributes. In \cite{zhang2015embedding}, multiple labels are leveraged to inject hierarchical inter-class relationship of attributes into learning feature representation . Lin \textit{et al.} \cite{zhou2015fine} utilized bipartite-graph labels to model rich inter-class relationships based on multiple sub-categories, which can be elegantly incorporated into convolutional neural network. However, those methods focus on the inter-class similarity distance, whereas the intra-class variance and its related triplet embedding have not been well studied in learning feature representation. When a category exhibits high intra-class appearance variance, intra-class triplet embedding is useful to deal with the complexity of feature space.
%Those intra-class attributes are effective to capture difference within a sub-category., while do not consider the intrinsic structure on intra-class attributes, the intrinsic structure on intra-class variance have not been well studied in learning robust fine-grained features. When a category exhibits high intra-class variance in appearance, they are difficult to cope with the complexity of data distribution.

\begin{figure*}[htb]
  \centering
  \vspace*{-8pt}
  \includegraphics[width=0.82\linewidth]{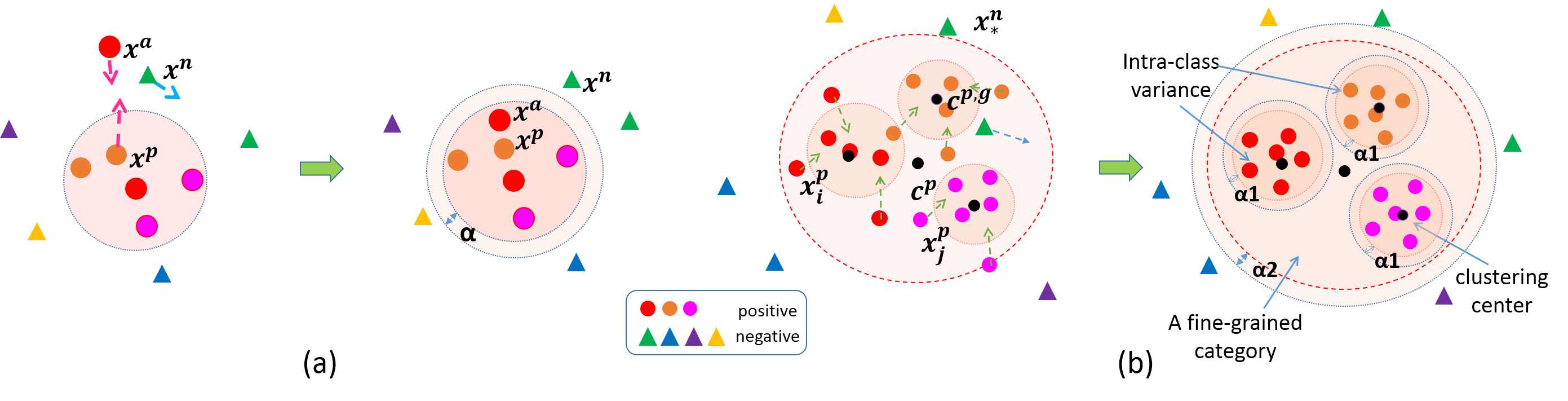}
  \vspace*{-15pt}
  \caption{Illustration of traditional triplet loss (a) and the intra-class variance (ICV) incorporated triplet loss (b). The instances denoted by different colors in (b), which can be sorted out by grouping in terms of some features or attributes. The ICV triplet loss further enforces that the samples within each group should be drawn closer. By contrast, the traditional triplet loss in (a) does not take the intra-class structure into account (Best viewed in color).}%Illustration of the intuition behind triplet loss and incorporating intra-class variance cluster triplet loss. Positives with different colors indicates have different intra-class attributes. %Negatives with different colors represent different categories.
  %Triplet loss constrains the same category samples into a local space, such that the intra-class variance structure has not been preserved. By contrast, proposed ICV triplet loss constrains the same categories together and meanwhile groups the samples with the same intra-class attributes closer.}
  \label{fig:triplet_intra-class_vision}
  \vspace*{-15pt}
\end{figure*}
\label{sec:format}
\label{triplet_loss}

In this paper, we propose a novel Group Sensitive TRiplet Sampling (GS-TRS) approach, which attempts to incorporate the modeling of intra-class variance into triplet network. A so-called grouping is to figure out a mid-level representation within each fine-grained category to capture the intra-class variance and intra-class invariance. In practice, clustering can be applied to implement the grouping.  Given a fine-grained category, instances are clustered to a set of groups. % different groups represent intra-class variance.% On the other hand, inter-class similarity can be represented by similarity between groups from different fine-grained categories. different fine-grained categories represent different
To formulate the triplet loss function, we need to consider the inter-class triplet embedding and the inter-group triplet embedding. The latter works on intra-class variance. The proposed GS-TRS has been proved to be effective in triplet learning, which can significantly improve the performance of triplet embedding in the presence of considerable intra-class variance.

Our main contributions are twofold. Firstly, we incorporate the modeling of intra-class variance into triplet network learning, which can significantly mitigate the negative impact of inter-class similarity and/or intra-class variance on fine-grained classification. Secondly, by optimizing the joint objective of softmax loss and triplet loss, we can generate effective feature representations (\textit{i.e.}, feature maps in Convolution Neural Network) for fine-grained retrieval. %Third, we propose a generic method to group instances within same class by means of clustering. % By exploiting the intra-class variance, higher priority can be assigned to the image with similar variance to the query image. Secondly, by combining triplet loss and softmax loss functions, the proposed method can not only get higher retrieval performance but also bring benefits to the classification as well. Thirdly, we propose a generic method to obtain intra-class variance label, which is realized by clustering method. %The proposed method is further evaluated on a Re-identification dataset\cite{liu2016deep} and a fine-grained dataset\cite{yang2015large}.
In extensive experiments over benchmark, the proposed method outperforms state-of-the-art fine-grained visual recognition approaches.

The rest of this paper is organized as follows. In Section 2, we formulate the problem of injecting the modeling of intra-class variance into triplet embedding for fine-grained visual recognition.  In Section 3, we present the proposed GS-TRS approach. Extensive experiments are discussed in Section 4, and finally we conclude this paper in Section 5.

\section{Problem Statement}

\subsection{Problem Formulation}

Let $S_{c,g}$ denote a set of instances of the $g_{th}$ group in fine-grained category $c$, and $S_n$ are a set of instances not in category $c$. Assume each category $c$ consists of $G$ groups, where the set of distinct groups may represent intra-class variance, and each individual group may represent intra-class invariance. The objective of preserving intra-class structure in metric  learning is to minimize the distances of samples in the same group for each category when the distances of samples from different categories exceed a minimum margin $\alpha$.
%Here we are particularly interested in maintaining intra-class variance through adding multi-level relationship to triplet.
\begin{equation}
\begin{aligned}
\vspace*{-25pt}
%\resizebox {1\hsize}{!}{${\min{\sum_{x_i, x_j \in S_{c,g}} {\|x_i-x_j\|^2}}  \quad s.t. \quad \max{\sum_{x_i, x_j \in S_{c,g}}{\|x_i-x_j\|^2 }} \le \alpha}.$}
\resizebox{0.6\hsize}{!}{${\min{ \sum_{g=1}^{G} \sum_{x_i, x_j \in S_{c,g}} {\|x_i-x_j\|^2}}}$} \quad  \\
\resizebox{0.6\hsize}{!}{${s.t. \quad \sum_{x_i\in S_{c,g}, x_k \in S_{n}}{\|x_i-x_k\|^2 }}$}\ge \alpha,
\end{aligned}
\label{problemdefinition}
\end{equation}
where samples $x_i$ and $x_j$ from category $c$ fall in the same group $g$; $x_k$ is from the other category; and $\alpha$ is the minimum margin constraint between samples from different categories.

Eq (\ref{problemdefinition}) can be optimized by deep metric learning using triplet network. The remaining issue is to model the intra-class variance of each fine-grained category and properly establish triplet units to accommodate the variance structure.

 \vspace*{-10pt}
\subsection{Triplet Learning Network}
 \vspace*{-5pt}
Our proposed GS-TRS approach works on a triplet network model. The main idea of triplet network is to project images into a feature space where those pairs belonging to the same category are closer than those from different ones. Let $<x^a, x^p, x^n>$ denote a triplet unit, where $x^a$ and $x^p$ belong to the same category, and $x^n$ belongs to the other category. The constraint can be formulated as:
\begin{equation}
\setlength{\abovedisplayskip}{5pt}
\setlength{\belowdisplayskip}{5pt}
\|f(x^a)-f(x^p)\|^2+\alpha \le \|f(x^a)-f(x^n)\|^2,
\label{eq:triplet_constraint}
\end{equation}
where $f(x)$ is the feature representation of image $x$, $\alpha$ is the minimum margin between positives and negatives. If the distances between positive and negative pairs violate the constraint in \eqref{eq:triplet_constraint}, then loss will be back propagated. Thus, the loss function can be defined as:
\begin{equation}
\setlength{\abovedisplayskip}{5pt}
\setlength{\belowdisplayskip}{5pt}
\resizebox{1\hsize}{!}{${L = \sum^N \frac{1}{2}\text{max}\{\|f(x^a)-f(x^p)\|_2^2+\alpha -\|f(x^a)-f(x^n)\|_2^2, 0\}}$}.
\label{eq:triplet}
\end{equation}
However, there exist two practically important issues in triplet network. First, triplet loss constrains samples of the same class together, while the class-inherent relative distances associated with intra-class variance cannot be well preserved, as illustrated in Fig. \ref{fig:triplet_intra-class_vision} (a). Second, triplet loss is sensitive to the selection of anchor $x^a$, and improper anchors  can seriously degrade the performance of triplet network learning.

 \vspace*{-10pt}
\section{GS-TRS APPROACH}
 \vspace*{-5pt}
The proposed GS-TRS incorporates intra-class variance into triplet network in which the learning process involves: (1) clustering each category into groups, (2) incorporating intra-class variance into triplet loss, (3) a multiple loss function.

\begin{figure}
\centering
%\vspace*{-5pt}
  \setlength{\abovedisplayskip}{3pt}
\setlength{\belowdisplayskip}{3pt}
  \includegraphics[width=0.85\linewidth]{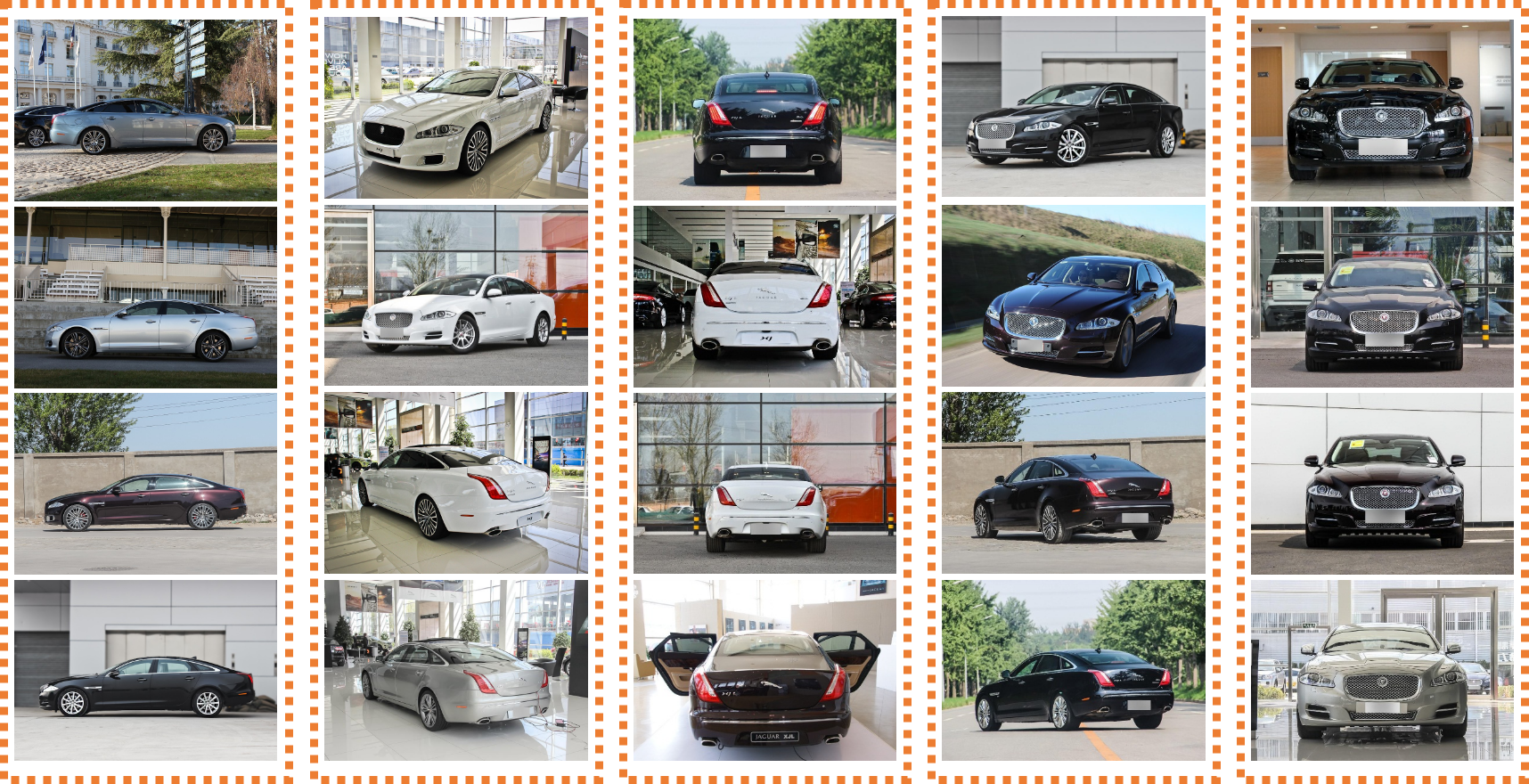}
  \vspace*{-5pt}
  \caption{ Exemplar car images from different groups, which are obtained by applying clustering ($K=5$) to the images of a specific car model in CompCar dataset. Different groups may be interpreted by some attributes (\textit{e.g.}, viewpoints or colors.) }%K-means clustering results ($K=5$) on samples within a specific car model in CompCar dataset, where each column denote a group with same or similar attributes such as viewpoints or colors.}
  \label{fig:kmeans.png}
  \vspace*{-15pt}
\end{figure}
\vspace*{-5pt}
\subsection{Intra-class Variance}
 \vspace*{-5pt}
To characterize intra-class variance, grouping is required. Unlike category labels, intrinsic attributes within a category are latent or difficult to precisely describe  (\textit{e.g.} lighting conditions, backgrounds). %Besides, supervised methods for obtaining intra-class variance have the high risk of over-fitting. In view of this,
Here, we prefer an unsupervised approach to grouping images for each category.%employ an unsupervised method to grouping the images in each category.
%Regarding to vehicles, different angles, light conditions and backgrounds are difficult to precisely describe.

Firstly, we feed image instances in each fine-grained category into the VGG\_CNN\_M\_1024 (\textit{VGGM}) network obtained by pre-training on ImageNet dataset. Then, we extract the last fully-connected layer's output as the feature representation, followed by Principal Component Analysis (PCA) based feature dimension reduction. Finally, K-means is applied to perform clustering:
\begin{equation}
\begin{array}{cl}
\setlength{\abovedisplayskip}{3pt}
\setlength{\belowdisplayskip}{6pt}
\arg{min}\sum_{g=1}^{G}\sum_{x=1 }^{N^{p,g}}\parallel f(x) - \mu_g\parallel^2,
\end{array}
\end{equation}
where $G$ is the number of cluster center $\mu_g$ (\textit{i.e.}, group num). $N^{p,g}$ is the number of samples contained in $S_{c,g}$.
Each image instance is assigned a group ID after clustering. As illustrated in Fig.~\ref{fig:kmeans.png}, grouping often relates to meaningful attributes.
%$S_{c,g}$ is the $g_{th}$ group under one category which represent a specific attribute.
%where $K$ is the cluster num of fine-grained category, and $I_k$ is the $k_{th}$ subset. %$C$ is the feature set consisting of last fully connected layer output $r$. Here $\mu_k$ indicates the cluster center.
%Those labels can be generated before training, such that the computational cost can be greatly reduced through one-time computation.

\subsection{Mean-valued Triplet Loss}
An anchor in triplet units is often randomly selected from positives. To alleviate the negative effects of improper anchor selection, we determine the anchor by computing the mean value of all positives, and formulate a mean-valued triplet loss.
%As illustrated in Eq(\ref{eq:triplet}), $x^a$ is anchor sample which belong to the same class as $x^p$. In practice the anchor $x^a$ is randomly selected from the samples belonging to the same class as $x^p$.
%However, as discussed in Sec.2 the selection of anchor has a great influence on learning stage.
Given a positive set $X^p=\{x^p_1,\cdots, x^p_{N^p}\} $ containing $N^p$ positive samples and a negative set $X^n=\{x^n_1,\cdots, x^n_{N^n}\} $ containing $N^n$ samples from other categories. Thus, the mean-valued anchor can be formulated as:
\begin{equation}
\setlength{\abovedisplayskip}{6pt}
\setlength{\belowdisplayskip}{6pt}
\resizebox{0.4\hsize}{!}{${c^p = \frac{1}{N^p}\sum_i^{N^p}f(x_i^p)}$},
\end{equation}
where $1\le i \le N^p$ and $1\le j \le N^n$. Rather than using randomly selected anchors, the proposed mean-valued triplet loss function is formulated as follows:
\begin{equation}
\begin{array}{cl}
\setlength{\abovedisplayskip}{6pt}
\setlength{\belowdisplayskip}{2pt}
L(c^p, X^p, X^n)=\\
\sum_i^{N^p}\frac{1}{2}max\{\|f(x^p_i)-c^p\|^2_2+\alpha-\|f(x^n_*)-c^p\|_2^2,0\},
\end{array}
\end{equation}
where $x^n_*$ is the negative closest to anchor $c^p$. It is worthy to note that, although the mean value of positives is considered as an anchor, the backward propagation needs to get all the positives involved. The advantage will be demonstrated in the subsequent experiments. %Liu \textit{et al.} \cite{liu2016deep} also compute the mean value of positives as anchor, while in back propagation stage does not consider that all the positives should be propagated.
When the anchor is computed by all of the positives, the triplet $<c^p, x_i^p, x_j^n>$ may not satisfy the constraints $\|f(x_i^p)-c^p\|_2^2+\alpha \le \|f(x_j^n)-c^p\|^2$.
%$\forall 1\le i\le N^p \text{ and } 1\le j\le N^n$,
Hence, all the positives involving mean value computing are enforced to perform backward propagation. The partial derivative of positive sample $x_i^p$ is:
\begin{equation}
\setlength{\abovedisplayskip}{5pt}
\setlength{\belowdisplayskip}{5pt}
\resizebox{0.8\hsize}{!}{${\frac{\partial L}{\partial f(x_i^p)} = f(x^p_i)-c^p + \frac{1}{N^p}(f(x_*^n)-f(x_i^p))}$}.
\end{equation}
The partial derivative of other positives $x_k^p$ ($k!=i$) is:
\begin{equation}
\setlength{\abovedisplayskip}{5pt}
\setlength{\belowdisplayskip}{5pt}
\resizebox{0.5\hsize}{!}{${\frac{\partial L}{\partial f(x_k^p)} = \frac{1}{N^p}(f(x_*^n)-f(x_i^p))}$}.
\end{equation}
The partial derivative of negative samples is:
\begin{equation}
\setlength{\abovedisplayskip}{5pt}
\setlength{\belowdisplayskip}{5pt}
\resizebox{0.4\hsize}{!}{${\frac{\partial L}{\partial f(x^n_*)} = c^p-f(x_*^n)}$}.
\end{equation}

%Compared to random triplet selection, online cluster triplet loss can better expresses the relationship between inter-class and intra-class, and also boost the convergence.

%It is significant to constrain the samples of the same category together in the meanwhile the intra-class relative distance can be well preserved.
%As illustrated in Eq(\ref{})
%Thus we adopt two groups of triplet to construct the loss function, and the intra-class variance maintaining loss (ICVM) is defined as follows:
%\begin{equation}
%\begin{array}{ll}
%L_{ICVM}=L_{triplet}(a_i^p, x_i^p, x^n) + L_{triplet}(a_i^p, x_j^p, x^n)\\[3mm]
%\begin{array}{cl}
%\resizebox {.9\hsize}{!}{$=\sum_i^{N^p}\frac{1}{2}max\{0, \|f(x_i^p)-c_i^p\|^2_2+\alpha_1-\|f(x^n)-c_i^p\|^2_2\}$}\\[2mm]
%\resizebox {.9\hsize}{!}{$+\sum_i^{N^p}\frac{1}{2}max\{0, \|f(x_j^p)-c_i^p\|^2_2+\alpha_2-\|f(x^n)-c_i^p\|^2_2\}.$}
%\end{array}
%\end{array}
%\end{equation}
%
%In this manner, our method is easy to be incorporated into the existing neural network.%our approach not only learns the discriminative feature between classes, but also maintains and utilizes the intra-class variance. Moreover,
%\subsection{Multi-loss Learning Framework}
%fine grained feature representation is faced with a problem: (2) The performance of triplet network is susceptible to triplet selection., since it can not cover the total triplet units $N^3$ in the training set due to the limited batch size.

\begin{figure*}[tb]
  \centering
  \vspace*{-8pt}
  \includegraphics[width=0.82\linewidth]{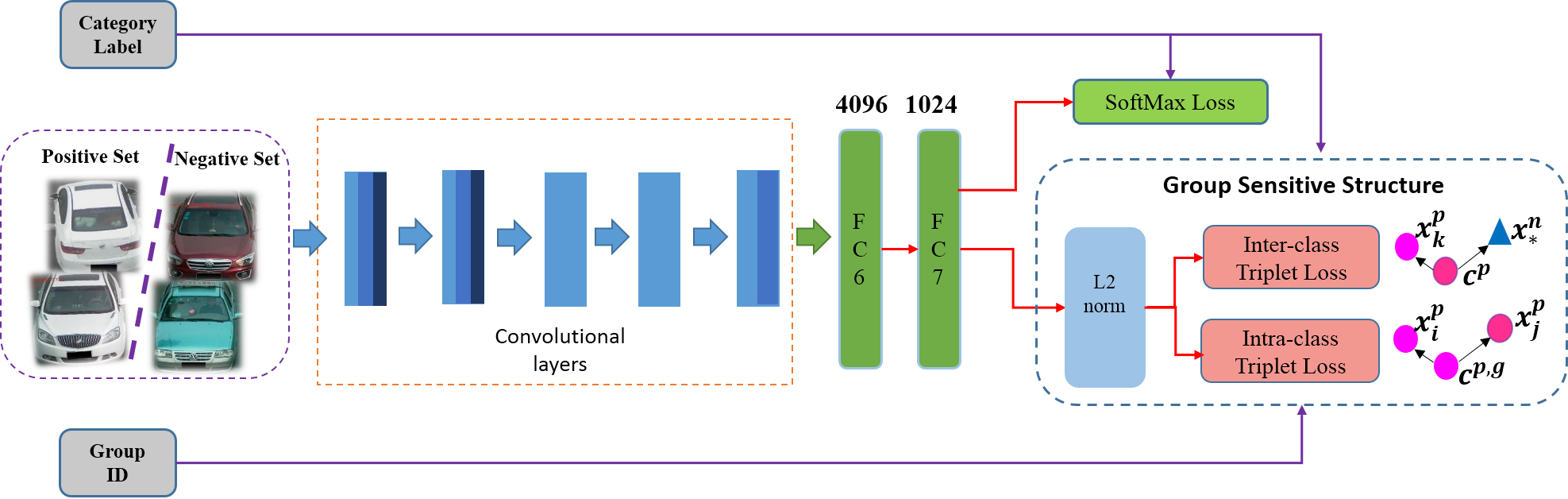}
  \vspace*{-5pt}
  \caption{Illustration of a triplet network by incorporating intra-class variance into triplet embedding, in which the joint learning objective is to minimize the combination of softmax loss and triplet loss (consisting of inter-class and intra-class triplet loss). }
  \vspace*{-5pt}
  %Intra-class variance structure network. Input minibatch contains groups of positive and negative samples set, that all calculated softmax loss and ICV  triplet loss. ICV  triplet loss consists of inter-class triplet loss and intra-class triplet loss embedded in the network structure.}
  \label{fig:Intra_mining_network}
  \vspace*{-10pt}
\end{figure*}
\subsection{Incorporating Intra-Class Variance into Mean-valued Triplet Loss}
To enforce the preservation of relative distances associated with intra-class variance, we introduces Intra-Class Variance loss (ICV loss) into triplet learning.
%can not present that the instances with the same or similar intra-class attributes are close to each other within the same category. In view of this,
%We consider to incorporate the intra-class variance into triplet learning (ICV triplet loss), which means to present a ``similar attribute, closer distance'' distribution of samples in one category.
Let $c^p$ denote a mean center (the mean value of samples) in category $c$ and $c^{p,g}$ denote a group center that is the mean value of samples in group $g$ of category $c$. For each category $c$, there are one mean center $c^p$ and $G$ group centers $c^{p,g}$. As illustrated in Fig. \ref{fig:triplet_intra-class_vision} (b), each black dot represents the center of a group. %Group division can be done according to intra-class variance discussed in Sec.3.1. $N^p$ and $N^{p,g}$ are the total number of samples in $c$ and $g_{th}$ group in $c$, respectively.  In Fig. \ref{fig:triplet_intra-class_vision} (b), for intra-class relationship,

In terms of intra-class variance, $x^p_i, x^p_j$ denote two samples from different groups within $c$. In terms of inter-class relationship, $x^p_k \in c$ are positives, and $x_*^n \notin c$ are  negatives. %I, $x^p_k$ represent all samples in $c$ (large circle) and $x^p_i,x^p_j$ represent the samples in different $g$ (small circle).
To incorporate the intra-class variance into triplet embedding, we formulate the constraints as:
\begin{equation}
\begin{aligned}
\setlength{\abovedisplayskip}{5pt}
\setlength{\belowdisplayskip}{6pt}
\resizebox{0.7\hsize}{!}{${\|c^{p}-f(x^p_i)\|^2+\alpha_1 \le \|c^{p}-f(x^n_*)\|^2 }$}\\
\resizebox{0.7\hsize}{!}{${\|c^{p,g}-f(x^p_i)\|^2+\alpha_2 \le \|c^{p,g}-f(x^p_j)\|^2 },$}
\end{aligned}
\end{equation}
where $\alpha_1$ is the minimum margin between those samples from different categories, and $\alpha_2$ is the minimum margin between those samples from different groups within the same category. Accordingly, we formulate the ICV incorporated mean-valued triplet loss as follows: %Therefore, the loss function is the combination of inter and intra-class variance that can be formulated as:
\begin{normalsize}
\vspace*{-5pt}
\begin{equation}
\footnotesize
\begin{aligned}
%&L_{total} = L_{inter}+L_{intra} \\
%\setlength{\abovedisplayskip}{3pt}
%\setlength{\belowdisplayskip}{0pt}
&L_{ICV\_Triplet} = L_{inter}(c^p,x_k^p,x_*^n) + {\!\sum_{g=1}^{G}}L_{intra}(c^{p,g},x_i^p,x_j^p) \\
&=\sum_{k=1}^{N_p} \frac{1}{2} \max{\{\|c^p - f(x_k^p)\|^2\!+\!\alpha_1\!-\!\|c^p-f(x_*^n)\|^2,0}\label{eq:total}\} \\
&{\! +\sum_{g=1}^{G}\!\sum_{\substack{i=1}}^{N^{p,g}} \! \frac{1}{2}\!  \max{\{\|c^{p,g}\!- \!f(x_i^p)\|^2\!+\!\alpha_2\!-\!\|c^{p,g}\!-\!f(x_j^p)\|^2,0}\}}.
\end{aligned}
\vspace*{-15pt}
\end{equation}
\end{normalsize}
\subsection{Joint Optimization of Multiple Loss Function}
ICV triplet loss alone does not suffice for effective and efficient feature learning in triplet network. Firstly, given a dataset of N images, the number of triplet units is $O(N^3)$, while each iteration in training often selects dozens of triplet units, and only a minority may violate the constraints. So the solely ICV triplet loss based learning incurs much slower convergence than classification. Secondly, as the triplet loss works on similarity distance learning rather than hyperplane decision, the discriminative ability of features can be improved by adding the classification loss to the learning objective. Hence, we propose a GS-TRS loss to jointly optimize the ICV triplet combinatin loss and softmax loss in a multi-task learning manner. A simple linear weighting is applied to construct the final loss function as follows:
%Solely using ICV triplet loss for feature learning has several demerits. For instance, given a dataset with $N$ images, the total number of triplet units can be $N^3$. Each iteration in training stage can only select dozens of triplet units, and only some of them may violate the constraints which may lead to slow convergence compared to classification. Moreover, triplet learning focuses on similarity comparisons not on hyperplane dicisions, therefore the distinguishing ability of learned feature can be inferior to using softmax loss (classification loss function) on classification task.  Thus, we propose to jointly optimize combo loss and softmax loss by a multitask learning method. The final loss function is integrated with these two losses through a weighted combination:(\textit{e.g.}, softmax loss)
%\textbf{Joint optimization } As mentioned before, the feature extracted from pure metric learning framework might not be suitable for classification task. To address that issue, we utilize softmax to assist intra-class variance mining. Softmax is widely applied in classification that is capable of projecting the samples of same class into a corner in feature space. In practice, we loose the strong softmax constraint to build an approximate classification feature distribution, and under this circumstance, intra-class variance mining can further refine this relative separation and obtain more stable performance. Thus we integrate them through a weighted sum fashion:
\begin{equation}
\setlength{\abovedisplayskip}{6pt}
\setlength{\belowdisplayskip}{6pt}
L_{GS-TRS} = \omega L_{softmax}+ (1-\omega )L_{ICV\_trplet},
\label{eq:total}
%\vspace*{-5pt}
\end{equation}
where $\omega$ is fusion weight. Fig.3 illustrates the triplet network. Optimizing this multi-loss function helps accomplish promising fine-grained categorization performance as well as discriminative features for fine-grained retrieval. We will investigate the effects of ICV\_triplet loss with or without mean-valued anchor on GS-TRS loss in the experiments.
%Fig.~\ref{fig:Intra_mining_network} visualizes network structure of our proposed method.  Simultaneously optimizing multi loss can not only learn discriminative features but also preserves an intra-class variance structure, which can benefit to the performance on both classification and retrieval tasks.

% In contrast to \cite{zhang2015embedding} \cite{cui2015fine} which adopted triplet network with three branches and computed the softmax loss of the anchor samples only, our network has only one data branch such that the entire mini-batch samples can contribute to both softmax and ICVM loss. Furthermore, the online hard triplet selection can be performed flexibly in our framework.

 \vspace*{-10pt}
\section{Experiments}
\label{sec:typestyle}
\subsection{Experiments Setup}

\noindent\textbf{Baselines} %The comparison of four methods that generate fine-grained feature representation are performed.
To evaluate and compare the triplet network based fine-grained visual recognition methods, we setup baseline methods as follows:
(1) triplet loss \cite{schroff2015facenet}, (2) triplet + softmax loss \cite{wang2014learning}, (3) mixed Diff + CCL \cite{liu2016deep}, (4) HDC + Contrastive \cite{yuan2016hard}, (5) GS-TRS loss without a mean-valued anchor for each group, \textit{i.e.}, a randomly selected anchor (GS-TRS loss W/O mean), (6) GS-TRS loss with a mean-valued anchor for each group (GS-TRS loss W/ mean). %(4) intra-class variance incorporated triplet loss + softmax loss(GS-TRS W/O cluster), (5) intra-class variance incorporated cluster-wise triplet loss + softmax loss (GS-TRS W/ cluster). %We follow the specifications from these compared papers for their settings and parameters.
%Regarding the hyper parameters, we set $\alpha=0.4$ in triplet, and $\alpha_1=0.5$, $\alpha_2=0.4$ in ICVM. Cluster numbers in CompCar and VehicleID are set to be 5 and 2, respectively.
We select the output of L2 Normalization layer as feature representation for retrieval and re-identification (ReID) tasks. For fair comparison, we adopt the base network structure VGG\_CNN\_M\_1024 (\textit{VGGM}) as in \cite{liu2016deep}. The networks are initialized with the pre-trained model over ImageNet.
%For retrieval and re-identification tasks, we extract the outputs of $L2$ Normalization layer as feature representation. To ensure fair comparisons, base network structure we adopt is VGG\_CNN\_M\_1024 (\textit{VGGM})as in \cite{liu2016deep}.  All of these networks are initialized with models pretrained on Imagenet dataset.

\noindent\textbf{DataSet} Comparison experiments are carried out over benchmark datasets VehicleID \cite{liu2016deep} and CompCar \cite{yang2015large}. VehicleID dataset consists of 221,763 images with 26,267 vehicles (about 250 vehicle models) captured by different surveillance cameras in a city. There are 110,178 images available for model training and three gallery test sets. The numbers of gallery images in small, medium and large sets are 800, 1,600 and 2,400 for retrieval and re-identification experiments. CompCar is another large-scale vehicle image dataset, in which car images are mostly collected from Internet. We select the Part-I subset for training that contains 431 car models (16, 016 images) and the remaining 14,939 images for test. Note that all the selected images involve more or less backgrounds. We conduct retrieval and ReID experiments on VehicleID dataset, and retrieval and classification experiments on CompCar dataset. %VehicleID dataset consists of 221763 images with 26267 vehicles (about 250 vehicle models) captured by different surveillance cameras in a city. There are 110178 images for model training and three different scale test sets for evaluations. %The gallery images contained in small, medium and large test sets are 800, 1600 and 2400 respectively. The corresponding probe images in three test sets are 6532, 11395 and17638 respectively.Both of have a large collection of vehicle images designed for different tasks.
%CompCar is a recently released large-scale and comprehensive vehicle image database, where the contained images are mostly collected from Internet. We select Part-I subset that contains 431 car models with a total of 16016 images for training and 14939 images for evaluations.% In all experiments, the chosen images all contain backgrounds that are more challenging.
%In particular, we aims to demonstrate that the feature generated by our method can well preserve the intra-class variance that can obtain higher retrieval performance than other methods. Additionally, we also report the improvements on classification task.

\noindent\textbf{Evaluation Metrics}
For retrieval performance evaluations, we use mAP and mean precision @\textit{K}. For ReID evaluation, we apply the widely used cumulative match curve (CMC). For classification evaluation, we use the mean percentage of those images accurately classified as the groundtruth.

\begin{figure}[h]
\centering
\vspace*{-5pt}
\subfigure[GS-TRS Loss (ICV triplet+ softmax loss) ]{
\label{fig:subfig:a} %% 第一幅图的标签
\includegraphics[width=0.95\linewidth]{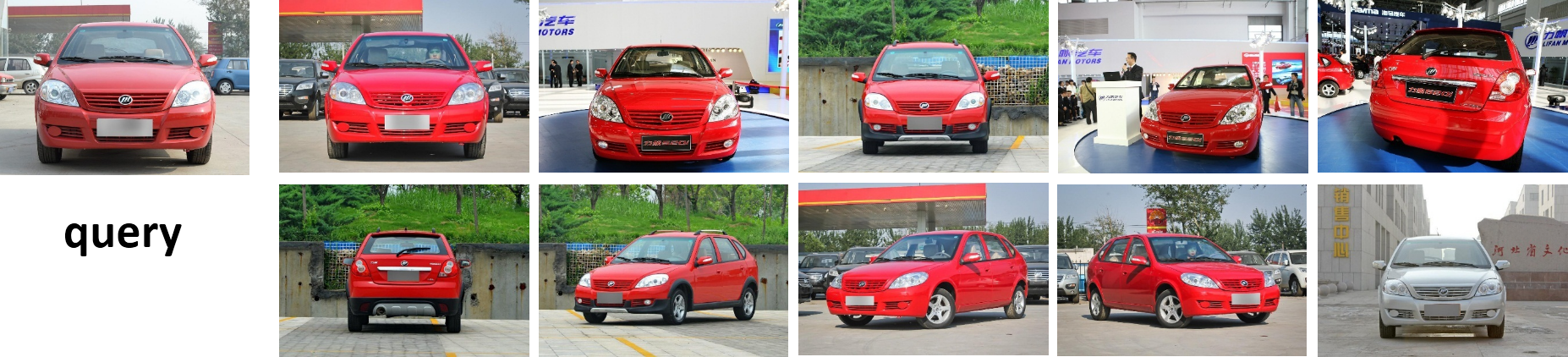}}
\hspace{1in} \subfigure[triplet loss + softmax loss]{
\label{fig:subfig:b} %% 第二幅图的标签
\includegraphics[width=0.95\linewidth]{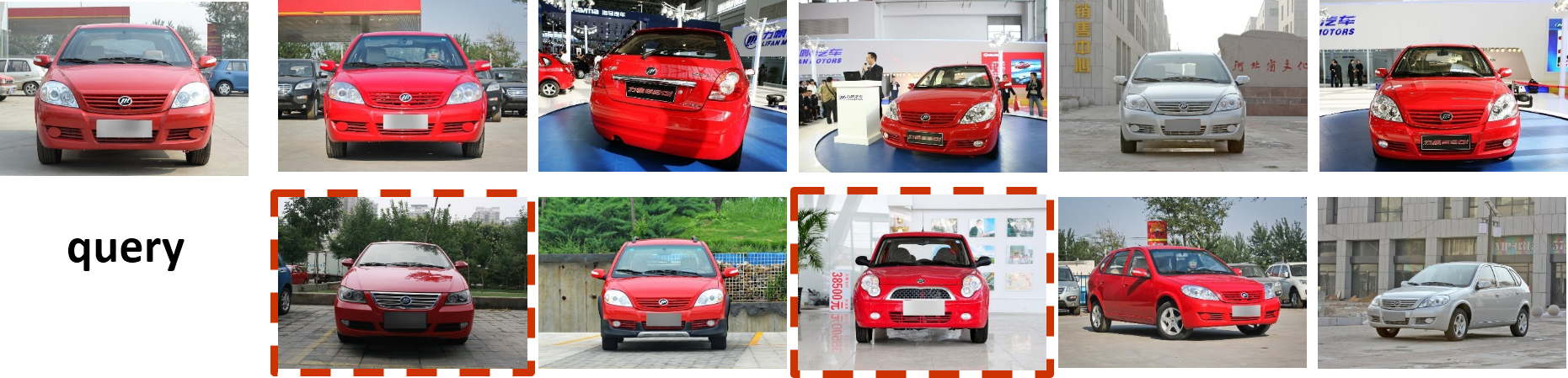}}
\vspace*{-8pt}
\caption{Exemplar Top-10 retrieval results on CompCar dataset. The images with a dashed rectangle are wrong results. The GS-TRS loss with grouping yields better results in (a) than the traditional triplet loss without grouping in (b).}%Visualization of Top 10 retrieval results on VehicleID Dataset. The images with dashed line are the wrong results. Recall images owning similar attributes with query can rank higher in (a) than in (b)}
\label{fig:retrieval.png}
\vspace*{-18pt}
\end{figure}

\subsection{Performance Comparison on VehicleID Dataset}
\label{sec:VehicleID}
%VehicleID dataset consists of 221763 images with 26267 vehicles (about 250 vehicle models) captured by different surveillance cameras in a city. There are 110178 images for model training and three different scale test sets for evaluations. The gallery images contained in small, medium and large test sets are 800, 1600 and 2400 respectively for retrieval and re-identification evaluations.

%Such more difficult classification objective can help network learn more discriminative features.

\textbf{Retrieval} Table \ref{tab:vehicle_retrieval} lists the retrieval performance comparisons. Note that during the training stage, unlike \cite{ding2015deep, liu2016deep} treating each vehicle model as a category, we treat each vehicle ID as a class (\textit{i.e.}, 13,134 vehicles classes). As listed in Table \ref{tab:vehicle_retrieval}, directly combining softmax and triplet loss has outperformed Mixed Diff+CCL \cite{liu2016deep} with significant mAP gain of 19.5\% in the large test set. Furthermore, our proposed GS-TRS loss without mean-valued anchors can consistently achieve significant improvements across three different scale subsets. In particular, the additional improvement on large test set reaches up to $4.6\%$ mAP. Compared to \cite{liu2016deep}, the improvement on large set has been up to 23.9\% mAP.  Moreover, GS-TRS loss with mean-valued anchors can further obtain about 2\% mAP gains since using mean values of positives from multiple groups within a category yields more reliable anchors, which contributes to better triplet embedding.

\begin{table}[h]\small
\vspace*{-15pt}
\caption{The mAP results of vehicle retrieval task.}
\vspace*{5pt}
\label{tab:vehicle_retrieval}
\centering
\begin{tabular}{|c|c|c|c|}
\hline
Methods & Small & Medium & Large\\
\hline
Triplet Loss~\cite{ding2015deep} & 0.444 & 0.391 & 0.373\\
CCL~\cite{liu2016deep} & 0.492 & 0.448 & 0.386\\
Mixed Diff+CCL~\cite{liu2016deep} & 0.546 & 0.481 & 0.455\\
Softmax Loss& 0.625 & 0.609 & 0.580\\
HDC + Contrastive~\cite{yuan2016hard} & 0.655 & 0.631 & 0.575\\
Triplet+Softmax Loss~\cite{wang2014learning}& 0.695 & 0.674 & 0.650\\
{GS-TRS loss W/O mean }& {0.731} & {0.718} & {0.696}\\
{GS-TRS loss W/ mean}& \textbf{0.746} & \textbf{0.734} & \textbf{0.715}\\
%{ICV Triplet+Softmax}& {0.731} & {0.718} & {0.696}\\
%{ICV Cluster Triplet+Softmax}& \textbf{0.746} & \textbf{0.734} & \textbf{0.715}\\
\hline
\end{tabular}
\vspace*{-5pt}
\end{table}

\begin{figure}[tbp]
  \centering
  \includegraphics[width=0.8\linewidth]{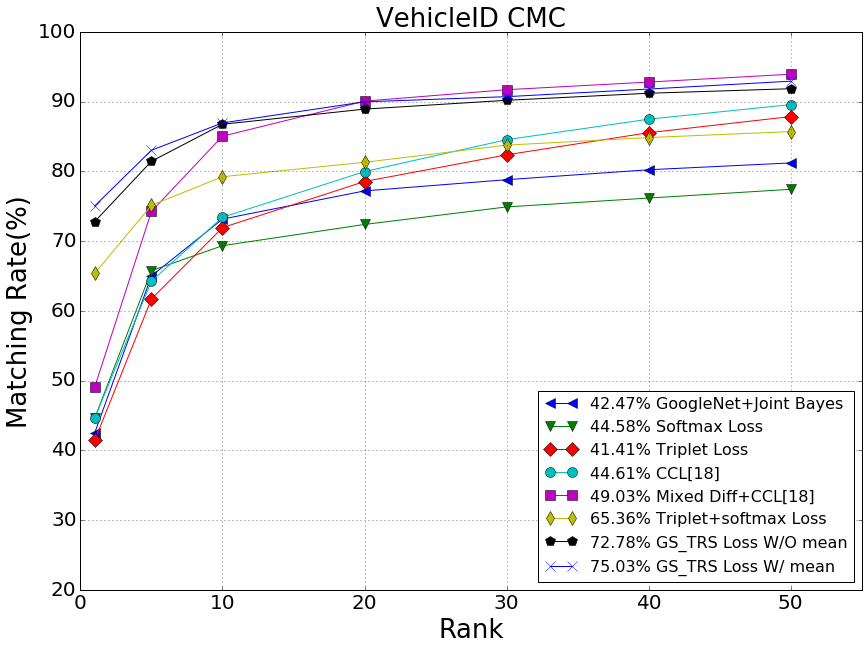}
  \vspace*{-10pt}
  \caption{CMC Results on VehicleID dataset.}
  \label{fig:cmc}
 \vspace*{-18pt}
\end{figure}

\textbf{Re-identification} Table \ref{tab:cmc_top_1_5} presents re-identification performance comparisons. %There are three different scale test sets gallery images contained in small, medium and large test sets are 800, 1600 and 2400 respectively. The corresponding probe images in three test sets are 6532, 11395 and 17638.
Our proposed method GS-TRS loss with mean-valued anchors achieves +30\% improvements over Mixed Diff+CCL in the large test set. Such significant improvements can be attributed to two aspects: First, we extend the softmax classification to the granularity level of vehicle ID, rather than the granularity level of vehicle model in \cite{liu2016deep}. Second, we have improved the similarity distance learning by introducing the intra-class feature space structure and its relevant loss function to triplet embedding. Moreover, from the performance comparisons of combining different triplet loss functions and softmax loss in Top1 and Top5, both the proposed GS-TRS loss without mean-valued anchors and the further improved GS-TRS loss with mean-valued anchors have yielded significant performance gains. More match rate details of different methods from Top 1 to Top 50 on the small test set are given in Fig.~\ref{fig:cmc}. %The detailed match rate from top-1 to top-50 of the various models evaluated on the small-scale test data is illustrated in Fig.~\ref{fig:cmc}.
%Compared with other methods, our approach presents significant advantageous on before top10 and comparable performance on top50.  %These results provide strong evidence that intra-class variance plays an important role in fine-grained similarity comparison.
%can significantly enhance the similarity representation that is pretty useful in retrieval and ReID tasks.
% MAP
\begin{table}[h]\footnotesize
\vspace*{-10pt}
\caption{The results of match rate in vehicle ReID task.}
\vspace*{5pt}
\label{tab:cmc_top_1_5}
\centering
%\begin{tabular}{|p{3.8cm}|p{0.5cm}|p{0.7cm}|p{0.9cm}|p{0.7cm}|}
\begin{tabular}{|c|c|c|c|c|}
\hline
\multicolumn{2}{|c|}{Method} & Small & Medium & Large\\
\hline\centering
Triplet Loss~\cite{ding2015deep} & \multirow{6}{*}{Top 1} & 0.404 & 0.354 & 0.319\\\centering
CCL~\cite{liu2016deep} & & 0.436 & 0.370 & 0.329\\\centering
Mixed Diff+CCL~\cite{liu2016deep} & & 0.490 & 0.428 & 0.382\\\centering
Triplet+Softmax Loss~\cite{wang2014learning}& & 0.683 & 0.674 & 0.653\\\centering
{GS-TRS loss W/O mean}& & {0.728} & {0.720} & {0.705}\\\centering
{GS-TRS loss W/ mean}& & \textbf{0.750} & \textbf{0.741} & \textbf{0.732}\\
\hline\centering
Triplet Loss~\cite{ding2015deep} & \multirow{6}{*}{Top 5} & 0.617 & 0.546 & 0.503\\\centering
CCL~\cite{liu2016deep} & & 0.642 & 0.571 & 0.533\\\centering
Mixed Diff+CCL~\cite{liu2016deep} & & 0.735 & 0.668 & 0.616\\\centering
Triplet+Softmax Loss~\cite{wang2014learning}& & 0.771 & 0.765 & 0.751\\\centering
{GS-TRS loss W/O mean}& & {0.814} & {0.805} & {0.789}\\\centering
{GS-TRS loss W/ mean}& & \textbf{0.830} & \textbf{0.826} & \textbf{0.819}\\
\hline
\end{tabular}
\vspace*{-20pt}
\end{table}

\subsection{Performance Comparison on CompCar Dataset}
%CompCar is a recently released large-scale and comprehensive vehicle image database, where the contained images are mostly collected from Internet. We select Part-I subset that contains 431 car models with a total of 16016 images for training and 14939 images for evaluations. In all experiments, the chosen images all contain backgrounds that are more challenging.

\textbf{Retrieval} Table \ref{tab:compcar_retrieval} lists the TopK precision comparisons. The incorporation of intra-class variance into triplet embedding can achieve more than $5.6\%$ precision gains at top-500. Overall, the modeling of intra-class variance and its injection into triplet network can significantly improve the discriminative power of feature representation which plays a significant role in fine-grained image retrieval. Fig.~\ref{fig:retrieval.png} gives the retrieval results of an  exemplar query over CompCar dataset before and after injecting GS-TRS into triplet embedding. %plays a significant for fine-grained image retrieval. It is worth mentioning that  through incorporating the intra-class variance, recall images with the same color as the query can acquire higher ranks in the recall list, since colors are typical intra-class variances, as shown in Fig.~\ref{fig:pic/retrieval.png}.

%\textbf{Re-identification.} In Fig.~\ref{fig:cmc}, the single triplet loss only achieves $17.58\%$ match rate at top-1, while our method brings $21.5\%$ improvements. Compared with multi-loss, our proposed ICVM method achieves $2.2\%$ higher precision at top-1 match rate and $7.5\%$ higher at top-50 match rate, which reveal the significant advantages of the proposed method.
% CMC
%\begin{figure}
%\begin{minipage}[b]{.48\linewidth}
%  \centering
%  \centerline{\includegraphics[width=4.0cm]{pic/cmc_compcars_intra.png}}
%  %\centerline{CMC on CompCars Dataset.}\medskip
%\end{minipage}
%\hfill
%\begin{minipage}[b]{.48\linewidth}
%  \centering
%  \centerline{\includegraphics[width=4.1cm]{pic/cmc_vehicle_intra.png}}
%  %\centerline{CMC on CompCars Dataset.}\medskip
%\end{minipage}
%
%\caption{CMC on CompCars dataset and VehicleID dataset.}
%\label{fig:cmc}
%\end{figure}
%% MAP
%\begin{table}[h]\small
%\vspace*{-15pt}
%\caption{Top K precision on CompCars retrieval task.}
%\vspace*{5pt}
%% \setlength{\abovecaptionskip}{0.cm}
%\label{tab:compcar_retrieval}
%\begin{tabular}{p{3.8cm}p{0.45cm}p{0.45cm}p{0.45cm}p{0.45cm}p{0.45cm}}
%\hline\centering
%Top K precision & 1 & 5 & 50 & 500 & mAP\\
%\hline\centering
%Triplet Loss\cite{ding2015deep} & 0.502 & 0.487 & 0.371 & 0.198 & 0.122\\\centering
%Softmax Loss& 0.456 &0.425 & 0.282 &  0.167 & 0.091\\\centering
%Triplet+Softmax Loss& 0.719 & 0.711 & 0.586 & 0.419 & 0.349\\\centering
%{ICV Triplet+Softmax}& 0.734 & 0.725 &0.603 & 0.475 & 0.376\\\centering
%{ICV Cluster Triplet+Softmax}& \textbf{0.756} & \textbf{0.733} &\textbf{0.620} & \textbf{0.497} & \textbf{0.393}\\
%\hline
%\end{tabular}
%\vspace*{-5pt}
%\end{table}

% MAP
\begin{table}[h]\footnotesize
\vspace*{-10pt}
\caption{mean precision @ \textit{K} on CompCars retrieval task.}
\vspace*{5pt}
\label{tab:compcar_retrieval}
\begin{tabular}{|c|c|c|c|c|}
\hline\centering
mean precision @ \textit{K} & 1 & 50 & 500 & All (mAP)\\
\hline\centering
Triplet Loss~\cite{ding2015deep} & 0.502 & 0.371 & 0.198 & 0.122\\\centering
Softmax Loss& 0.456  & 0.282 &  0.167 & 0.091\\\centering
Triplet+Softmax Loss~\cite{wang2014learning}& 0.719 & 0.586 & 0.419 & 0.349\\\centering
{GS-TRS loss W/O mean}& 0.734 &0.603 & 0.475 & 0.376\\\centering
{GS-TRS loss W/ mean}& \textbf{0.756} &\textbf{0.620} & \textbf{0.497} & \textbf{0.393}\\
\hline
\end{tabular}
\vspace*{-5pt}
\end{table}

\noindent\textbf{Classification} We train a \textit{VGGM} network with single softmax loss and set initial learning rate = 0.002 and total iteration = 80K, and then yield $78.24\%$ classification accuracy. Further fine-tuning with triplet+softmax loss can bring about $0.7\%$ classification accuracy improvement, while GS-TRS loss with mean-valued anchors can yield more accuracy improvement of $1.6\%$ (\textit{i.e.}, the classification accuracy is 79.85\%). Such improvements demonstrate that preserving intra-class variance is beneficial for fine-grained categorization as well.
 \vspace*{-8pt}
\section{Conclusion}
\label{sec:majhead}
 \vspace*{-5pt}
We have proposed a novel approach GS-TRS to improve triplet network learning through incorporating the intra-class variance structure into triplet embedding. The multi-task learning of both GS-TRS triplet loss and softmax loss has significantly contributed to fine-grained image retrieval and classification. How to further optimize the grouping strategy as well as  the selection of anchors with respect to meaningful and effective groups is included in our future work.
%In this paper, we have proposed a learning framework to effectively generate fine-grained feature representation by incorporating intra-class variance. In our method, intra-class variance is incorporated into proposed cluster triplet loss, which consider adding different constraints according to the intra-class attributes similarity in feature representation. The proposed method considerably improved the image retrieval precision on two fine-grained datasets. In the future, how to make full use of metric information to further improve the classification performance will be studied.

\vspace*{10pt}
\textbf{Acknowledgments:} This work was supported by grants from National Natural Science Foundation of China (U1611461, 61661146005, 61390515) and National Hightech R\&D Program of China (2015AA016302). This research is partially supported by the PKU-NTU Joint Research Institute, that is sponsored by a donation from the Ng Teng Fong Charitable Foundation.

\bibliographystyle{IEEEbib}
\bibliography{icme2017template}

\end{document}